\documentclass[journal]{IEEEtran}
\usepackage{graphicx}
\usepackage{amsmath}
\usepackage{float}
\usepackage{multirow}
\usepackage{booktabs}
\usepackage{amssymb}
\usepackage{cite}
\usepackage[hidelinks]{hyperref}

\title{DGCM-SSFA: Dual-Granularity Cross-Modulation and Sparse Strip Frequency Attention for Efficient 3T-to-7T MRI Synthesis}
\author{Zhichun Fan, Zhiqiang Chen, Lanling Liu, Lingjun Liu, Zhonghua Xie$^{*}$\\
{\small School of Computer Science and Engineering, Huizhou University, Huizhou, China}\\
\thanks{$^{*}$Corresponding author: eexie@qq.com. This work was supported by National Natural Science Foundation of China (62001184), the Projects of Educational Commission of Guangdong Province (2025ZDZX1038, 2025KTSCX49, 2023ZDZX1025), Graduate Education Innovation Program of Guangdong Province (2026JGXM\_064), Huizhou University Independent Innovation Capability Enhancement Plan Project (HZU202535), College Students' Innovative Entrepreneurial Training Plan Program (202410577008, 202610577034).}}
\date{June 2026}

\begin{document}

\maketitle

\begin{abstract}
Synthesizing high-fidelity 7T MRI from routinely acquired 3T scans offers a practical pathway to enhanced neuroimaging without the prohibitive cost of ultra-high-field scanners. While existing deep learning approaches have achieved promising results, state-of-the-art methods typically rely on either large-scale architectures exceeding 45 million parameters or computationally intensive ensemble strategies, limiting their practicality in resource-constrained clinical environments. In this work, we propose a compact multi-domain attention framework, termed DGCM-SSFA, that achieves competitive synthesis quality with only 11M parameters. The proposed architecture integrates two complementary modules into a U-Net Transformer backbone. First, a Dual-Granularity Cross-Modulated Feature Modulator (DGCM) is embedded at multiple hierarchical levels to enhance local edge and texture details through Sobel-guided spatial attention and depthwise frequency-aware channel attention with cross-branch gating. Second, a Sparse Strip Frequency Attention (SSFA) module at the bottleneck captures global spatial context via strip-pooled attention with near-linear complexity, while simultaneously performing FFT-based spectral modulation in the frequency domain. These two modules operate synergistically with the backbone's native channel attention, forming a triple-granularity feature refinement pipeline. Extensive experiments on the UNC and BNU 3T-to-7T MRI datasets across T1w and T2w modalities demonstrate that DGCM-SSFA achieves strong structural fidelity, attaining an SSIM of 0.8431 on BNU T2w---approaching the best benchmark value of 0.8624---and 0.8100 on BNU T1w, while using substantially fewer parameters than Transformer-based alternatives such as ResViT (123.4M) and FS-RWKV (45.4M). These results validate that carefully designed multi-domain attention provides a practical and parameter-efficient solution for clinical MRI synthesis.
\end{abstract}

\begin{IEEEkeywords}
MRI synthesis, 3T-to-7T translation, multi-domain attention,
frequency-aware modulation, sparse strip attention, compact network,
medical image synthesis
\end{IEEEkeywords}

\section{Introduction}
Magnetic resonance imaging at 7 Tesla (7T) represents the current gold standard for structural neuroimaging, offering an order-of-magnitude gain in signal-to-noise ratio and spatial resolution over conventional 3T scanners. This elevated tissue contrast enables the delineation of subcortical nuclei, cortical microarchitecture, and subtle white matter lesions that remain invisible at lower field strengths. Despite these compelling diagnostic advantages, 7T adoption is severely limited by hardware costs exceeding several million dollars, specialized site requirements, and stringent safety protocols---factors that collectively restrict access to a handful of elite research centers worldwide \cite{7tmri}.

Synthesizing 7T-equivalent images from routine 3T acquisitions has therefore attracted substantial interest \cite{fsrwkv,chen2026image,11098517}. The task is to learn a mapping that recovers the high-resolution anatomical details latent in clinical scans, yet this mapping must capture fine-grained edge structures at tissue boundaries, long-range anatomical dependencies across entire brain regions, and subtle textural variations, which remains an open challenge.

Early approaches to this problem employed deep convolutional architectures \cite{unet,ddcnn,chen2026docunfold} and generative adversarial formulations \cite{cyclegan,pix2pix,reggan}. These methods exhibit a characteristic trade-off: CNNs faithfully reconstruct local intensity patterns but fail to model dependencies beyond their limited receptive field, whereas GANs can hallucinate visually convincing textures at the cost of introducing spurious anatomical features that undermine diagnostic trustworthiness. More recent Transformer-based designs \cite{vit,swin,DBLP:conf/cvpr/ShiTHJCPWP26} address the global context deficit through self-attention, but their quadratic complexity forces aggressive downsampling that discards precisely the high-frequency details most critical for clinical interpretation. The latest generation of state space models \cite{mamba,vmamba} partially resolves this tension by offering linear-complexity global modeling; however, they operate predominantly in the spatial domain and provide no explicit mechanism for frequency-selective feature refinement.

A deeper issue cuts across these paradigms. Anatomical structures in MRI span multiple scales and exhibit distinct signatures in both the spatial and frequency domains. Edge contours at tissue boundaries are best captured by local gradient operators, global intensity distributions across an entire slice are naturally modeled by long-range attention, and fine periodic textures are most compactly represented in the Fourier domain. Yet no existing cross-field synthesis framework integrates all three forms of attention---local edge-guided, global spatial, and frequency-domain---within a single unified architecture. This motivates the central premise of our work: that a carefully orchestrated synergy of complementary attention mechanisms can achieve competitive synthesis fidelity without the massive parameter counts of monolithic Transformer or state space models.

We propose DGCM-SSFA, a compact multi-domain attention network built upon a U-Net Transformer backbone. Its first component (DGCM) is deployed at multiple encoder levels and combines an edge-aware spatial branch with a frequency-aware channel branch; the two branches gate each other \cite{gated_scnn}. The second component (SSFA) operates at the bottleneck, compressing the key-value token space from $H\times W$ to $H+W$ for near-linear global attention \cite{strip_pool}, with a parallel FFT branch that performs spectral modulation of the magnitude component. Both modules are detailed in Section~III.

The principal contributions of this work are:
\begin{enumerate}
    \item A multi-domain attention architecture, DGCM-SSFA, that integrates dual-granularity cross-modulated feature enhancement and sparse strip frequency attention into a U-Net Transformer for 3T-to-7T MRI synthesis, operating at 11M parameters.
    \item The DGCM module, which unifies Sobel-structured edge priors, depthwise spectral estimation, and spatial-channel gating to amplify local anatomical detail.
    \item The SSFA module, which achieves global context modeling with near-linear complexity through strip-pooled attention and incorporates FFT-based spectral modulation for frequency-aware refinement.
    \item Extensive experiments on the UNC and BNU datasets across T1w and T2w modalities, demonstrating that DGCM-SSFA achieves strong structural fidelity---attaining an SSIM of 0.8431 on BNU T2w, approaching the best benchmark value of 0.8624 while using only 11M parameters, substantially fewer than Transformer-based alternatives.
\end{enumerate}

\section{RELATED WORK}
\subsection{Deep Learning for Cross-Field MRI Synthesis}

Mapping 3T MRI to its 7T counterpart lies at the intersection of super-resolution and cross-modal translation. Early approaches used convolutional architectures trained with pixel-wise regression \cite{unet,ddcnn}, achieving reasonable intensity fidelity but producing over-smoothed outputs that lacked crisp anatomical boundaries. Adversarial training \cite{cyclegan,pix2pix,reggan} substantially improved perceptual quality, but introduces a well-documented risk of hallucinating spurious anatomical features \cite{reggan,cyclegan}---a dangerous failure mode in diagnostic imaging.

Transformer-based architectures \cite{resvit,vit,swin} address the smoothness issue through self-attention, capturing long-range dependencies that CNNs miss, but ResViT \cite{resvit} combines residual convolutional blocks with vision transformers at the cost of 123M parameters. State space formulations \cite{fsrwkv,mamba,vmamba} offer a more efficient alternative, with FS-RWKV \cite{fsrwkv} showing that linear-complexity recurrent architectures can approach Transformer-level quality. Despite these advances, the dominant design philosophy remains one of scaling, leaving a gap for compact architectures that achieve competitive quality through structural innovation.

\subsection{Attention Mechanisms in Image Restoration}

Channel attention, as popularized by SENet \cite{senet} and refined by CBAM \cite{cbam}, recalibrates feature responses by modeling interdependencies between channels through global pooling operations. Spatial attention \cite{nonlocal} complements channel attention by generating a 2D weight map that highlights informative regions; however, standard spatial attention based on max- and average-pooling statistics lacks sensitivity to structural edges, motivating the use of explicit gradient operators such as Sobel filters \cite{sobel} to inject edge-awareness.

A parallel line of work explores frequency-domain processing for image restoration \cite{fft_restore,dwt_restore,focal_freq,chen2026aquaclarity}. Different degradation types manifest in distinct frequency bands: high-frequency components carry edge and texture information, while low-frequency components encode global illumination and contrast. Fast Fourier Convolution \cite{fft_restore} performs convolution in the complex frequency domain through a dual-branch design, whereas DCT-based learning \cite{focal_freq} applies channel selection for efficient spectral processing.

For global context aggregation, strip pooling \cite{strip_pool} offers an attractive compromise between the limited receptive field of local operations and the quadratic cost of dense self-attention by compressing the attention token space from $O(HW)$ to $O(H+W)$. This technique is increasingly adopted in restoration tasks where global illumination and contrast consistency are critical.

\subsection{Multi-Domain and Multi-Granularity Feature Fusion}

A recurring theme in recent restoration literature is combining complementary attention mechanisms. CBAM \cite{cbam} pioneered this idea by stacking channel and spatial attention, and subsequent work extended it to the frequency domain \cite{focal_freq,dwt_restore}. Others use cross-modal gating \cite{gated_scnn,xu2026mage}, where one branch modulates another. Despite these advances, existing MRI synthesis frameworks typically employ at most two attention types and rarely integrate frequency-domain processing or edge-aware spatial attention within a unified design. Our work addresses this gap by orchestrating three complementary attention mechanisms---local edge-guided spatial, global strip-pooled spatial, and frequency-domain spectral---coupled through cross-branch and cross-domain gating.

\section{METHOD}

\subsection{Overall Pipeline}

\begin{figure*}[t]
    \centering
    \includegraphics[width=\textwidth]{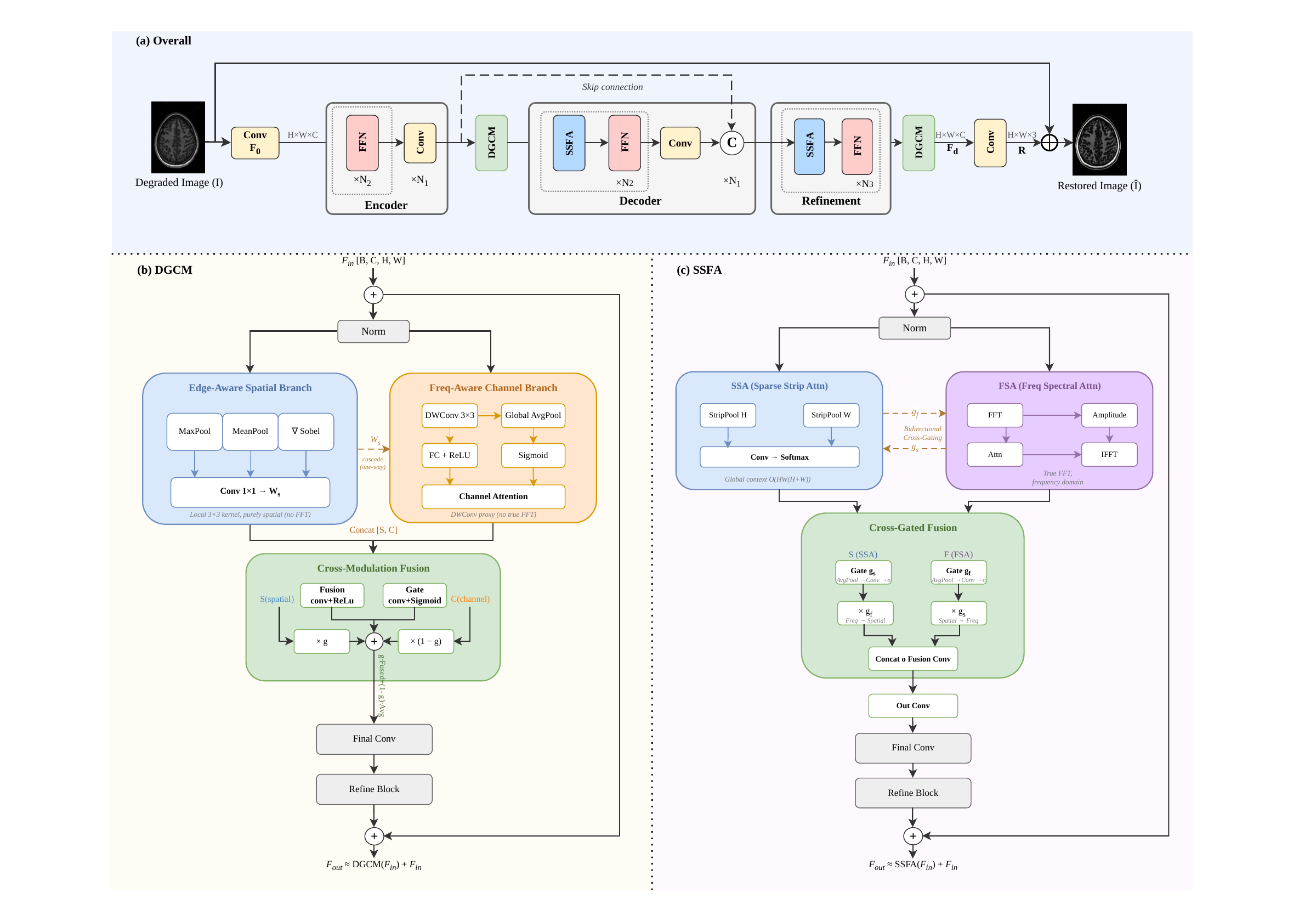}
    \caption{Overall architecture of the proposed DGCM-SSFA. (a) The framework adopts a U-Net-style encoder-decoder backbone. The encoder stacks FFN and Conv layers for hierarchical feature extraction. DGCM modules are inserted after each encoder level and after the refinement stage for local detail enhancement. A single SSFA module is deployed at the bottleneck for global spatial-frequency attention. A global residual connection produces the final output. (b) DGCM consists of an edge-aware spatial branch and a frequency-aware channel branch with cross-modulation fusion. (c) SSFA combines sparse strip attention (SSA) and frequency spectral attention (FSA) with cross-gated fusion.}
    \label{fig:overall}
\end{figure*}

The overall architecture of DGCM-SSFA is illustrated in 
Fig.~\ref{fig:overall}(a). Given a degraded 3T MRI image 
$I \in \mathbb{R}^{H \times W \times C}$, a convolutional 
layer first extracts shallow features 
$F_0 \in \mathbb{R}^{H \times W \times C}$. The encoder 
then progressively downsamples the feature maps through a 
stack of FFN modules and convolutional layers, capturing 
hierarchical representations at multiple scales. A DGCM 
module is inserted after each encoder level to perform 
dual-granularity cross-modulated feature refinement, 
enhancing local edge and texture details at progressively 
deeper semantic levels.

At the bottleneck, the most compact latent representation is 
processed by the SSFA module, which captures global spatial 
context via strip-pooled attention and performs FFT-based 
spectral modulation in the frequency domain. The decoder 
symmetrically upsamples the features back to the original 
resolution through Transformer blocks with self-attention. 
A skip connection from the encoder output is concatenated 
with the decoder features to preserve low-level structural 
information.

The decoded features are further processed by a refinement 
stage consisting of additional Transformer blocks, 
followed by a fourth DGCM for final detail enhancement. The 
refined features are projected to the residual image 
$R \in \mathbb{R}^{H \times W \times C}$ via a convolutional 
layer. The final restored 7T-like MRI image is obtained 
through a global residual connection: 
$\hat{I} = I + R$.

The encoder and decoder consist of repeated Transformer blocks, each with
a Multi-head Self-Attention (MSA) module and a Feed-Forward Network (FFN)
applying a pointwise convolution with GELU activation. The
encoder downsamples via stride-2 convolution across three
levels, while the decoder symmetrically upsamples through
bilinear interpolation. Skip connections concatenate low-level
encoder features with high-level decoder features, preserving
fine-grained spatial information lost during compression.

The two core modules, DGCM and SSFA, are detailed in 
Fig.~\ref{fig:overall}(b) and (c), respectively. DGCM 
employs an edge-aware spatial branch and a frequency-aware 
channel branch with cross-modulation gating to enhance local 
feature discrimination. SSFA combines sparse strip attention 
with FFT-based spectral attention under a cross-gated fusion 
mechanism for efficient global context modeling. These two 
modules operate in a complementary manner: DGCM focuses on 
local edge and texture refinement at multiple hierarchical 
levels, while SSFA captures long-range dependencies at the 
bottleneck.

\subsection{DGCM: Dual-Granularity Cross-Modulated Feature Modulator}

The DGCM module, illustrated in Fig.~\ref{fig:overall}(b), is deployed at four
hierarchical positions: after the first, second, and third encoder levels, and
after the refinement stage. At each position, DGCM takes a feature map
$F_{\text{in}} \in \mathbb{R}^{C \times H \times W}$ and produces a refined
output $F_{\text{out}}$ of identical dimensions, following a local-global
residual design:
\begin{equation}
F_{\text{out}} = F_{\text{in}} + \mathcal{R}(F_{\text{fused}}),
\end{equation}
where $\mathcal{R}$ denotes a lightweight residual refinement block composed
of two convolutional layers with ReLU activation.

\textit{1) Edge-Aware Spatial Branch:} The spatial branch constructs a
three-channel descriptor from the input feature map: a channel-wise maximum map
highlighting strong activations, a channel-wise mean map capturing overall
intensity distribution, and a gradient magnitude map computed via fixed Sobel
operators applied to the mean map. The Sobel kernels $G_x$ and $G_y$ are
registered as non-trainable buffers, ensuring zero additional parameter cost for
edge extraction. The gradient magnitude is computed as:
\begin{equation}
G_{\text{mag}} = \sqrt{(G_x \ast F_{\text{mean}})^2 + (G_y \ast F_{\text{mean}})^2 + \epsilon},
\end{equation}
where $\ast$ denotes convolution and $\epsilon$ is a small constant for
numerical stability. The three-channel descriptor is passed through a lightweight
convolutional head consisting of two $3\times3$ convolutions with an intermediate
ReLU activation, producing a single-channel spatial attention map
$W_s \in [0,1]^{1 \times H \times W}$ via sigmoid activation. The spatially
modulated output is:
\begin{equation}
F_{\text{sp}} = W_s \odot F_{\text{in}},
\end{equation}
where $\odot$ denotes element-wise multiplication with broadcasting.

\textit{2) Frequency-Aware Channel Branch:} The channel branch first transforms
the input through two consecutive $1\times1$ convolutions to obtain an
intermediate representation $F_d$. Three statistical descriptors are then
computed: a max-pooled vector $z_{\text{max}} \in \mathbb{R}^{C \times 1 \times 1}$,
an average-pooled vector $z_{\text{avg}}$, and a frequency-aware vector
$z_{\text{freq}}$ obtained by applying a depthwise $3\times3$ convolution to
$F_d$, modulating the result with the spatial attention map $W_s$, and globally
average-pooling the absolute response:
\begin{equation}
z_{\text{freq}} = \operatorname{GAP}\bigl(| \operatorname{DWConv}_{3\times3}(F_d) \odot W_s |\bigr).
\end{equation}
The three descriptors are concatenated and processed by a two-layer MLP
with a reduction ratio of $4$, producing a channel attention vector
$W_c \in [0,1]^{C \times 1 \times 1}$.
The channel-modulated output is:
\begin{equation}
F_{\text{ch}} = W_c \odot F_{\text{in}}.
\end{equation}

\textit{3) Cross-Modulation Fusion:} Rather than simply summing the two branch
outputs, DGCM employs a learnable gated fusion mechanism. The spatial and
channel outputs are concatenated and processed by two parallel pathways: a
fusion pathway that generates an integrated feature map $F_{\text{conv}}$
through a $1\times1$--$3\times3$ convolution sequence, and a gating pathway
that produces a sigmoid-gated mask $G \in [0,1]^{C \times H \times W}$:
\begin{equation}
\begin{aligned}
F_{\text{fused}} = G \odot \operatorname{Conv}_{3\times3}\bigl(\operatorname{Conv}_{1\times1}([F_{\text{sp}}, F_{\text{ch}}])\bigr) \\
+ (1 - G) \odot \tfrac{1}{2}(F_{\text{sp}} + F_{\text{ch}}).
\end{aligned}
\end{equation}
This formulation allows the network to dynamically blend the fused representation
with the direct average of the two branches, providing a soft selection mechanism
that adapts to the local feature characteristics at each spatial position.

\subsection{SSFA: Sparse Strip Frequency Attention}

While DGCM excels at local detail enhancement, capturing long-range anatomical
dependencies---such as global intensity gradients across the brain, bilateral
symmetry, and large-scale tissue contrast---requires a global attention mechanism.
SSFA, deployed at the bottleneck (dimension $192$), addresses this through two
parallel pathways operating in the spatial and frequency domains, coupled by
cross-domain gating.

\textit{1) Sparse Strip Spatial Attention:} Standard dense self-attention incurs
$\mathcal{O}(H^2W^2)$ complexity, which is prohibitive for high-resolution
feature maps. SSFA adopts strip pooling to compress the key and value token
space from $HW$ to $H+W$. Given the query, key, and value projections
$Q, K, V \in \mathbb{R}^{B \times \text{head} \times HW \times d}$, the key and
value tensors are reshaped to 2D spatial form and average-pooled along each axis
independently:
\begin{equation}
K_h = \operatorname{AvgPool}_{W}(K_{2D}), \quad
K_w = \operatorname{AvgPool}_{H}(K_{2D}),
\end{equation}
\begin{equation}
K_{\text{strip}} = \operatorname{Concat}(K_h, K_w) \in \mathbb{R}^{B \times \text{head} \times (H+W) \times d},
\end{equation}
with analogous operations for $V$. The attention is then computed as:
\begin{equation}
\operatorname{Attn}_{\text{strip}} = \operatorname{Softmax}\!\left(\frac{Q K_{\text{strip}}^{\mathsf{T}}}{\sqrt{d}}\right) V_{\text{strip}},
\end{equation}
yielding $\mathcal{O}(HW(H+W))$ complexity---a substantial reduction from the
quadratic cost of dense attention, particularly at the bottleneck where spatial
dimensions are smallest.

\textit{2) Frequency Spectral Attention:} In parallel, the input feature map is
transformed to the frequency domain via the 2D real-valued Fast Fourier Transform
(rFFT). The resulting complex spectrum is decomposed into magnitude
$M = |\mathcal{F}(F)|$ and phase $\Phi = \angle\mathcal{F}(F)$. A channel
attention mechanism is applied exclusively to the magnitude component:
\begin{equation}
M' = M \odot \sigma\bigl(\operatorname{MLP}(\operatorname{GAP}(M))\bigr),
\end{equation}
where $\sigma$ denotes the sigmoid function and the MLP consists of two
$1\times1$ convolutions with a reduction ratio of $4$. The phase component is
preserved unchanged, as it encodes the structural positioning of features. The
refined spectrum is reconstructed via inverse rFFT:
$F_{\text{freq}}' = \mathcal{F}^{-1}(M' \cdot e^{j\Phi})$. This design enables the network to
selectively amplify or suppress specific frequency bands, enhancing
high-frequency tissue boundaries while preserving low-frequency contrast.

\textit{3) Cross-Domain Gating:} The spatial and frequency pathways are not
independent. A cross-domain gating mechanism enables mutual refinement: the
spatial output is gated by a channel-wise vector derived from the frequency
output, and vice versa:
\begin{equation}
F_{\text{sp}}' = F_{\text{sp}} \odot \sigma\bigl(\operatorname{Conv}_{1\times1}(\operatorname{GAP}(F_{\text{freq}}'))\bigr),
\end{equation}
\begin{equation}
F_{\text{freq}}'' = F_{\text{freq}}' \odot \sigma\bigl(\operatorname{Conv}_{1\times1}(\operatorname{GAP}(F_{\text{sp}}))\bigr).
\end{equation}
The gated outputs are concatenated and fused through a $1\times1$--$3\times3$
convolution sequence, followed by a final $1\times1$--$1\times1$ projection and
a global residual addition:
\begin{equation}
F_{\text{SSFA}} = F_{\text{in}} + F_{\text{proj}}.
\end{equation}

\subsection{Training Objective}

The model is trained with a simple L1 reconstruction loss between the synthesized
7T image $\hat{I}_{7T}$ and the ground truth $I_{7T}$:
\begin{equation}
\mathcal{L} = \frac{1}{N} \sum_{i=1}^{N} \bigl\| \hat{I}_{7T}^{(i)} - I_{7T}^{(i)} \bigr\|_1,
\end{equation}
where $N$ denotes the number of pixels. The L1 loss is chosen over L2 for its
robustness to outliers and its tendency to produce sharper outputs in image
restoration tasks~\cite{loss_l1}. The global residual formulation---where the
network predicts the difference between 3T and 7T rather than the 7T image
directly---further stabilizes training.

\subsection{Implementation Details}

DGCM-SSFA is implemented in PyTorch with the BasicSR
toolbox~\cite{basicsr}, and trained with AdamW~\cite{adamw} (initial learning rate
$3\times10^{-4}$, weight decay $1\times10^{-4}$) under a cosine annealing
schedule~\cite{cosine}. Training runs for 30,000 iterations with a two-stage
progressive patch size ($128\times128$ for the first 10,000 iterations, then
$256\times256$), a batch size of 8 per GPU on two NVIDIA RTX 5090 GPUs, and
standard geometric augmentation (random flipping, 90-degree rotation, and
cropping).

\section{EXPERIMENTS}

\subsection{Datasets and Preprocessing}

We evaluate on two publicly available paired 3T--7T MRI
datasets: UNC and BNU~\cite{fsrwkv}. UNC provides paired
3T (Siemens Prisma) and 7T (Siemens Terra) brain scans
from 10 participants (T1w: MPRAGE/MP2RAGE; T2w: SPACE/TSE);
BNU provides 20 subjects scanned on 3T Siemens Prisma and
7T Siemens Magnetom (T1w: MPRAGE/3D MPRAGE; T2w: TSE/3D
SPACE). Following standard preprocessing, all volumes are
resampled to isotropic 1.0~mm, skull-stripped, rigidly
registered across field strengths using FSL FLIRT~\cite{flirt},
and normalized to $[0,1]$. Each volume is sliced along the
axial plane into $256\times256$ 2D images.

\subsection{Evaluation Metrics}

We adopt three widely used quantitative metrics for image 
synthesis quality assessment: Peak Signal-to-Noise Ratio 
(PSNR), Structural Similarity Index Measure (SSIM)~\cite{ssim} 
computed with an $11\times11$ Gaussian kernel ($\sigma=1.5$), 
and Root Mean Square Error (RMSE). PSNR and RMSE capture 
pixel-level reconstruction fidelity, while SSIM reflects 
perceptual structural consistency. Higher PSNR and SSIM 
values indicate better performance, whereas lower RMSE 
denotes superior reconstruction accuracy.

\subsection{Comparison with State-of-the-Art Methods}

To comprehensively assess the synthesis capability of 
DGCM-SSFA, we compare it against eight representative 
methods spanning four architectural paradigms. The CNN-based 
category includes DDCNN~\cite{ddcnn} and WATNet~\cite{watnet}; 
GAN-based methods comprise CycleGAN~\cite{cyclegan}, 
pix2pix~\cite{pix2pix}, and RegGAN~\cite{reggan}; 
Transformer-based approaches cover ResViT~\cite{resvit} and 
PTNet~\cite{ptnet}; and the SSM-based baseline FS-RWKV~\cite{fsrwkv} 
represents the latest state-space model. All methods are trained 
and evaluated under a unified protocol on the same data splits.

\textbf{Quantitative Results.} Table~\ref{tab:comparison} 
reports the numerical comparison across four modality-site 
combinations. Our DGCM-SSFA achieves competitive performance 
with only 11M parameters. On the BNU T2w dataset, it attains 
the second-highest SSIM of 0.8431, trailing only FS-RWKV 
(0.8624), demonstrating the effectiveness of the multi-granularity 
attention design in preserving structural fidelity. On the BNU 
T1w dataset, the SSIM of 0.8100 ranks second only to FS-RWKV 
(0.8388). In terms of parameter efficiency, DGCM-SSFA contains 
11M parameters---significantly fewer than FS-RWKV (45.40M), 
ResViT (123.44M), pix2pix (57.18M), and PTNet (28.13M)---while 
maintaining competitive or superior structural similarity.

\begin{figure*}[t]
    \centering
    \includegraphics[width=\textwidth]{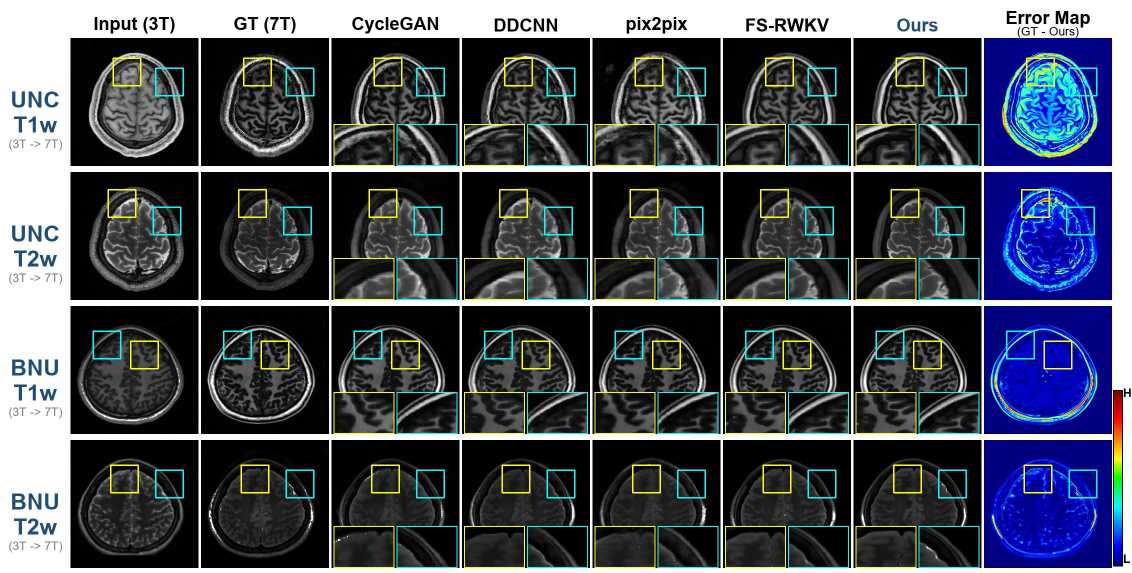}
    \caption{Qualitative comparison of synthesized 3T-to-7T MRI on four datasets (UNC T1w, UNC T2w, BNU T1w, BNU T2w). From left to right: 3T input, 7T ground truth, CycleGAN, DDCNN, pix2pix, FS-RWKV, the proposed DGCM-SSFA, and error map (error magnitude relative to GT). Yellow and cyan boxes in the main image denote regions magnified below.}
    \label{fig:visual}
\end{figure*}

\textbf{Qualitative Analysis.} Visual comparisons of 
synthesized 7T-like images are presented in 
Fig.~\ref{fig:visual}. DGCM-SSFA recovers sharper tissue boundaries and 
finer anatomical textures compared to competing methods, 
particularly in cortical gray matter and subcortical regions 
where high-frequency details are critical.

\begin{table*}[t]
\centering
\caption{Quantitative Comparison on MRI Cross-Modality Synthesis (3T $\rightarrow$ 7T)}
\label{tab:comparison}
\footnotesize
\setlength{\tabcolsep}{3pt}
\begin{tabular}{l ccc ccc ccc ccc c}
\toprule
Method & \multicolumn{3}{c}{UNC T1w} & \multicolumn{3}{c}{UNC T2w} & \multicolumn{3}{c}{BNU T1w} & \multicolumn{3}{c}{BNU T2w} & Params (M) \\
\cmidrule(lr){2-4} \cmidrule(lr){5-7} \cmidrule(lr){8-10} \cmidrule(lr){11-13}
& PSNR$\uparrow$ & SSIM$\uparrow$ & RMSE$\downarrow$ & PSNR$\uparrow$ & SSIM$\uparrow$ & RMSE$\downarrow$ & PSNR$\uparrow$ & SSIM$\uparrow$ & RMSE$\downarrow$ & PSNR$\uparrow$ & SSIM$\uparrow$ & RMSE$\downarrow$ & \\
\midrule
CycleGAN & 19.98 & 0.6789 & 0.1008 & 24.88 & 0.7538 & 0.0586 & 21.93 & 0.7881 & 0.0807 & 26.30 & 0.8053 & 0.0493 & 28.29 \\
pix2pix & 19.96 & 0.6847 & 0.1015 & 24.74 & 0.7550 & 0.0591 & 22.74 & 0.8102 & 0.0739 & 26.47 & 0.8182 & 0.0481 & 57.18 \\
DDCNN & 20.75 & 0.6412 & 0.0924 & 24.21 & 0.7286 & 0.0640 & 23.02 & 0.7732 & 0.0717 & 26.88 & 0.8229 & 0.0463 & 0.29 \\
WATNet & 20.17 & 0.6753 & 0.0990 & 23.07 & 0.6698 & 0.0719 & 21.13 & 0.7319 & 0.0886 & 26.13 & 0.7625 & 0.0503 & 1.35 \\
RegGAN & 19.46 & 0.6423 & 0.1073 & 23.97 & 0.6994 & 0.0654 & 22.10 & 0.7897 & 0.0793 & 25.61 & 0.7703 & 0.0531 & 2.76 \\
ResViT & 20.14 & 0.6696 & 0.0994 & 23.47 & 0.7140 & 0.0687 & 22.78 & 0.8069 & 0.0733 & 26.58 & 0.8251 & 0.0477 & 123.44 \\
PTNet & 20.82 & 0.7142 & 0.0916 & 24.66 & 0.7495 & 0.0602 & 22.98 & 0.7903 & 0.0717 & 27.34 & 0.8242 & 0.0439 & 28.13 \\
FS-RWKV & \underline{21.01} & \underline{0.7258} & \underline{0.0898} & \underline{25.31} & \underline{0.7807} & \underline{0.0565} & \underline{23.36} & \underline{0.8388} & \underline{0.0689} & \underline{27.49} & \underline{0.8624} & \underline{0.0431} & 45.40 \\
\midrule
Ours & 20.21 & \textbf{0.6924} & 0.0972 & 24.50 & \textbf{0.7505} & 0.0614 & 22.39 & \textbf{0.8100} & 0.0765 & 26.27 & \textbf{0.8431} & 0.0496 & 11 \\
\bottomrule
\end{tabular}
\end{table*}

\subsection{Ablation Studies}

\begin{figure}[htbp]
    \centering
    \includegraphics[width=\linewidth]{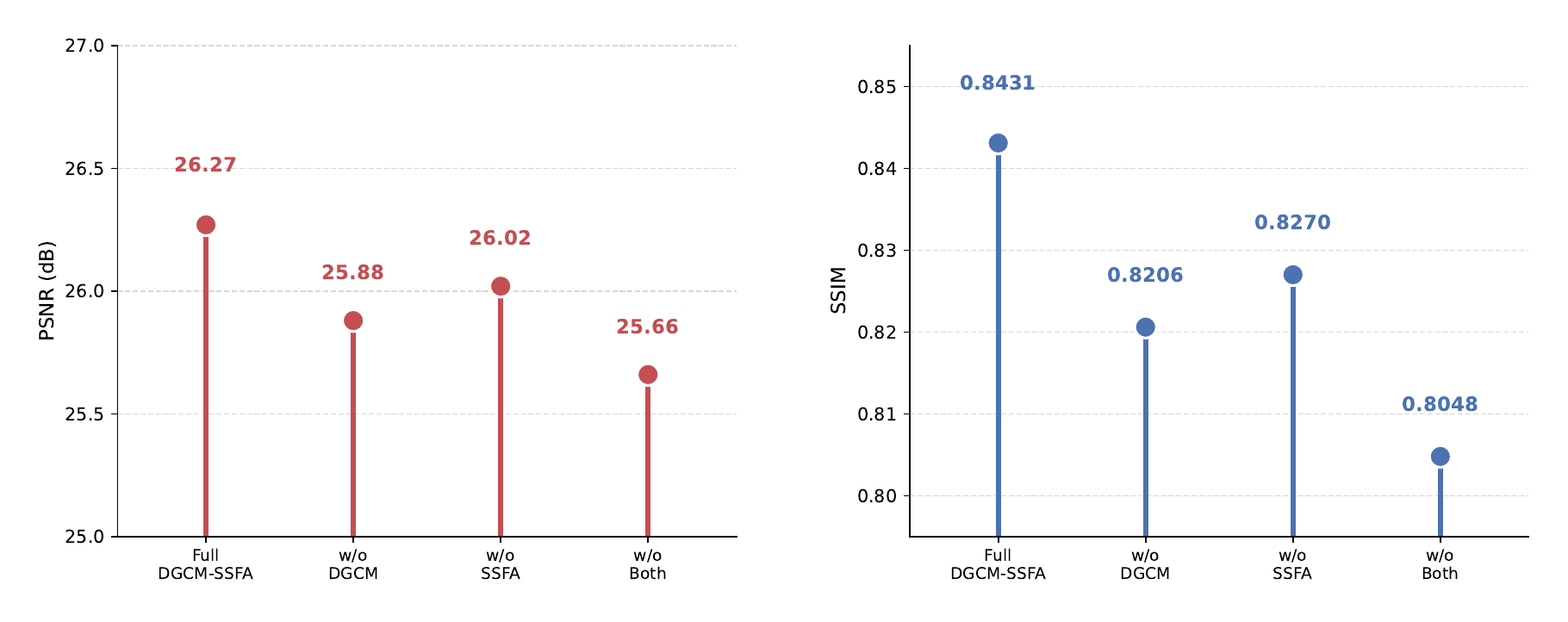}
    \caption{Ablation study on the key components of DGCM-SSFA. PSNR (left) and SSIM (right) are reported for four variants. All models are trained without ensemble and evaluated on BNU T2w.}
    \label{fig:ablation}
\end{figure}

We conduct ablation experiments on the BNU T2w dataset to validate the
contribution of each core component; quantitative results are summarized in
Table~\ref{tab:ablation} and visualized in Fig.~\ref{fig:ablation}.

\begin{table}[htbp]
\centering
\caption{Ablation study on the key components of DGCM-SSFA. All models are trained without ensemble and evaluated on BNU T2w. The best result in each column is highlighted in bold.}
\label{tab:ablation}
\begin{tabular}{lcc}
\toprule
Variant & PSNR$\uparrow$ & SSIM$\uparrow$ \\
\midrule
Full DGCM-SSFA          & \textbf{26.27} & \textbf{0.8431} \\
w/o DGCM                & 25.88 & 0.8206 \\
w/o SSFA                & 26.02 & 0.8270 \\
w/o Both                & 25.66 & 0.8048 \\
\bottomrule
\end{tabular}
\end{table}

Removing all DGCM modules yields a PSNR of 25.88~dB and SSIM of 0.8206,
degradations of 0.39~dB and 0.0225, confirming that the edge-aware spatial and
frequency-aware channel branches jointly preserve fine anatomical boundaries.
Removing SSFA gives 26.02~dB and 0.8270 (drops of 0.25~dB and 0.0161), with the
degradation most pronounced in regions of elongated anatomical structures,
validating the contribution of strip-pooled global attention and FFT-based
spectral modulation. With both modules removed, the backbone alone reaches
25.66~dB and 0.8048. The individual contributions ($0.39+0.25=0.64$~dB) closely
match the joint contribution (0.61~dB), leaving only 0.03~dB of functional
overlap and indicating that the two modules operate on largely orthogonal
feature spaces.

\section{DISCUSSION AND CONCLUSION}

DGCM-SSFA achieves competitive synthesis quality with only 11M parameters, showing that carefully designed multi-domain attention can rival larger architectures. On BNU T2w it attains the second-highest SSIM of 0.8431, approaching FS-RWKV (0.8624) while using only 24\% of its parameters. The ablation study further reveals that DGCM and SSFA operate on largely orthogonal feature spaces with only 0.03~dB of functional overlap. Despite strong structural fidelity, the model yields moderate PSNR, reflecting a preference for anatomical boundary preservation over pixel-wise intensity accuracy---a trade-off that is clinically acceptable given that diagnostic confidence depends primarily on the clear delineation of tissue interfaces. Several limitations merit future investigation: the 2D slice-wise formulation does not exploit inter-slice anatomical consistency, and a volumetric 3D extension could improve both fidelity metrics~\cite{11523523}; hybrid loss functions combining adversarial, perceptual~\cite{perceptual}, and frequency-domain objectives~\cite{chen2026image} may better balance structural and pixel-wise quality; and validation on pathological cases remains an important step toward clinical translation. In summary, DGCM-SSFA offers a practical and parameter-efficient solution for high-fidelity 3T-to-7T MRI synthesis.



\begin{thebibliography}{99}

\bibitem{7tmri}
S. M. Seyedsaadat, S. N. Seyed Saadat, V. Patel, S. J. S. Sandhu, D. Paech, A. Desai, X. Zhou, S. Tao, V. Gupta, and E. H. Middlebrooks,
``Ultrahigh-field 7-T MRI in neuroradiology: A comprehensive clinical review,'' 
\textit{RadioGraphics}, vol. 46, no. 1, e250194, 2026.

\bibitem{fsrwkv}
Y. Lei, Z. Li, Y. Liu, X. Chen, and C.-M. Pun,
``FS-RWKV: leveraging frequency spatial-aware RWKV for 3T-to-7T MRI translation,''
in \textit{Proc. IEEE Int. Conf. Bioinf. Biomed. (BIBM)}, 2025, pp. 1--6.

\bibitem{chen2026image}
X. Chen, C. Jing, Z. Li, Y. Shen, G. Huang, Y. Liu, B. Lei, C.-M. Pun, and S. Wang,
``From image synthesis to digital twins: A paradigm shift in AI-driven medical image computing,''
\textit{IEEE Trans. Consum. Electron.}, vol. 72, no. 3, pp. 6132--6144, 2026.

\bibitem{11098517}
X. Chen, Y. Shen, M. Mahmud, M. Kwok-Po Ng, K.-F. Tsang, S. Wang, C.-M. Pun, and S. Wang,
``ConnectomeDiffuser: Generative AI enables brain network construction from diffusion tensor imaging,''
\textit{IEEE Trans. Consum. Electron.}, vol. 71, no. 3, pp. 7835--7847, 2025.

\bibitem{ptnet}
X. Zhang, X. He, J. Guo, N. Ettehadi, N. Aw, D. Semanek, J. Posner, A. Laine, and Y. Wang,
``PTNet: a high-resolution infant MRI synthesizer based on transformer,'' 
arXiv:2105.13993, 2021.

\bibitem{resvit}
O. Dalmaz, M. Yurt, and T. \c{C}ukur,
``ResViT: residual vision transformers for multimodal medical image synthesis,''
\textit{IEEE Trans. Med. Imaging}, vol. 41, no. 10, pp. 2598--2614, 2022.

\bibitem{unet}
O. Ronneberger, P. Fischer, and T. Brox,
``U-Net: convolutional networks for biomedical image segmentation,''
in \textit{Proc. Int. Conf. Med. Image Comput. Comput.-Assist. Intervent. (MICCAI)}, 2015, pp. 234--241.

\bibitem{ddcnn}
K. Zhang, W. Zuo, Y. Chen, D. Meng, and L. Zhang,
``Beyond a Gaussian denoiser: residual learning of deep CNN for image denoising,''
\textit{IEEE Trans. Image Process.}, vol. 26, no. 7, pp. 3142--3155, 2017.

\bibitem{chen2026docunfold}
X. Chen, Z. Zhou, Z. Li, X. Zhang, Y. Dong, and K.-F. Tsang,
``DocUnfold: Leveraging unfolding network and a real-world large-scale dataset for handwriting contamination removal in documents,''
\textit{IEEE Trans. Consum. Electron.}, vol. 72, no. 1, pp. 1138--1145, 2026.

\bibitem{cyclegan}
J.-Y. Zhu, T. Park, P. Isola, and A. A. Efros,
``Unpaired image-to-image translation using cycle-consistent adversarial networks,''
in \textit{Proc. IEEE Int. Conf. Comput. Vis. (ICCV)}, 2017, pp. 2223--2232.

\bibitem{pix2pix}
P. Isola, J.-Y. Zhu, T. Zhou, and A. A. Efros,
``Image-to-image translation with conditional adversarial networks,''
in \textit{Proc. IEEE Conf. Comput. Vis. Pattern Recognit. (CVPR)}, 2017, pp. 1125--1134.

\bibitem{reggan}
L. Kong, C. Lian, D. Huang, Z. Li, Y. Hu, and Q. Zhou,
``Breaking the dilemma of medical image-to-image translation,''
in \textit{Proc. Adv. Neural Inf. Process. Syst. (NeurIPS)}, 2021, pp. 1964--1978.

\bibitem{vit}
A. Dosovitskiy, L. Beyer, A. Kolesnikov, D. Weissenborn, X. Zhai, T. Unterthiner, et al.,
``An image is worth 16x16 words: transformers for image recognition at scale,''
in \textit{Proc. Int. Conf. Learn. Represent. (ICLR)}, 2021.

\bibitem{swin}
Z. Liu, Y. Lin, Y. Cao, H. Hu, Y. Wei, Z. Zhang, et al.,
``Swin transformer: hierarchical vision transformer using shifted windows,''
in \textit{Proc. IEEE/CVF Int. Conf. Comput. Vis. (ICCV)}, 2021, pp. 10012--10022.

\bibitem{DBLP:conf/cvpr/ShiTHJCPWP26}
H. Shi, H. Tao, G. Huang, J. Jiang, X. Chen, C.-M. Pun, S. Wang, and P. Pan,
``Intrinsic concept extraction based on compositional interpretability,''
in \textit{Proc. IEEE/CVF Conf. Comput. Vis. Pattern Recognit. (CVPR)}, 2026, pp. 38969--38978.

\bibitem{mamba}
A. Gu and T. Dao,
``Mamba: linear-time sequence modeling with selective state spaces,''
in \textit{Proc. Conf. Lang. Model. (COLM)}, 2024.

\bibitem{vmamba}
Y. Liu, Y. Tian, Y. Zhao, H. Yu, L. Xie, Y. Wang, et al.,
``VMamba: visual state space model,''
in \textit{Proc. Adv. Neural Inf. Process. Syst. (NeurIPS)}, vol. 37, 2024.

\bibitem{gated_scnn}
T. Takikawa, D. Acuna, V. Jampani, and S. Fidler,
``Gated-SCNN: gated shape CNNs for semantic segmentation,''
in \textit{Proc. IEEE/CVF Int. Conf. Comput. Vis. (ICCV)}, 2019, pp. 5222--5231.

\bibitem{strip_pool}
Q. Hou, L. Zhang, M.-M. Cheng, and J. Feng,
``Strip pooling: rethinking spatial pooling for scene parsing,''
in \textit{Proc. IEEE/CVF Conf. Comput. Vis. Pattern Recognit. (CVPR)}, 2020, pp. 4003--4012.

\bibitem{senet}
J. Hu, L. Shen, and G. Sun,
``Squeeze-and-excitation networks,''
in \textit{Proc. IEEE Conf. Comput. Vis. Pattern Recognit. (CVPR)}, 2018, pp. 7132--7141.

\bibitem{cbam}
S. Woo, J. Park, J.-Y. Lee, and I. S. Kweon,
``CBAM: convolutional block attention module,''
in \textit{Proc. Eur. Conf. Comput. Vis. (ECCV)}, 2018, pp. 3--19.

\bibitem{nonlocal}
X. Wang, R. Girshick, A. Gupta, and K. He,
``Non-local neural networks,''
in \textit{Proc. IEEE Conf. Comput. Vis. Pattern Recognit. (CVPR)}, 2018, pp. 7794--7803.

\bibitem{sobel}
I. Sobel,
``An isotropic 3x3 image gradient operator,''
in \textit{Machine Vision for Three-Dimensional Scenes}, H. Freeman, Ed. Academic Press, 1990, pp. 376--379.

\bibitem{fft_restore}
L. Chi, B. Jiang, and Y. Mu,
``Fast Fourier convolution,''
in \textit{Proc. Adv. Neural Inf. Process. Syst. (NeurIPS)}, 2020, pp. 4479--4488.

\bibitem{dwt_restore}
H. Huang, R. He, Z. Sun, and T. Tan,
``Wavelet domain generative adversarial network for multi-scale face hallucination,''
\textit{Int. J. Comput. Vis.}, vol. 127, no. 6, pp. 763--784, 2019.

\bibitem{focal_freq}
K. Xu, M. Qin, F. Sun, Y. Wang, Y.-K. Chen, and F. Ren,
``Learning in the frequency domain,''
in \textit{Proc. IEEE/CVF Conf. Comput. Vis. Pattern Recognit. (CVPR)}, 2020, pp. 1740--1749.

\bibitem{chen2026aquaclarity}
X. Chen, Z. Li, X. Zhang, Y. Lei, Y. Dong, X. Zhang, and C.-M. Pun,
``AquaClarity: Vision transformer enhances underwater image via nonlinear feature fusion and color fidelity improvement,''
\textit{IEEE Trans. Consum. Electron.}, vol. 72, no. 1, pp. 1122--1129, 2026.

\bibitem{xu2026mage}
X. Xu, Y. Pan, X. Zhang, X. Chen, and K.-F. Tsang,
``Manifold-guided geometric enhancement for reliable multimodal interaction in consumer electronics,''
\textit{IEEE Trans. Consum. Electron.}, vol. 72, no. 1, pp. 1156--1163, 2026.

\bibitem{loss_l1}
H. Zhao, O. Gallo, I. Frosio, and J. Kautz,
``Loss functions for image restoration with neural networks,''
\textit{IEEE Trans. Comput. Imaging}, vol. 3, no. 1, pp. 47--57, 2017.

\bibitem{basicsr}
X. Wang, K. Yu, K. Zhou, W. Luo, and C. C. Loy,
``BasicSR: Open source image and video restoration toolbox,''
\url{https://github.com/XPixelGroup/BasicSR}, 2023.

\bibitem{adamw}
I. Loshchilov and F. Hutter,
``Decoupled weight decay regularization,''
in \textit{Proc. Int. Conf. Learn. Represent. (ICLR)}, 2019.

\bibitem{cosine}
I. Loshchilov and F. Hutter,
``SGDR: Stochastic gradient descent with warm restarts,''
in \textit{Proc. Int. Conf. Learn. Represent. (ICLR)}, 2017.

\bibitem{flirt}
M. Jenkinson, P. Bannister, M. Brady, and S. Smith,
``Improved optimization for the robust and accurate linear registration and motion correction of brain images,''
\textit{NeuroImage}, vol. 17, no. 2, pp. 825--841, 2002.

\bibitem{ssim}
Z. Wang, A. C. Bovik, H. R. Sheikh, and E. P. Simoncelli,
``Image quality assessment: from error visibility to structural similarity,''
\textit{IEEE Trans. Image Process.}, vol. 13, no. 4, pp. 600--612, 2004.

\bibitem{watnet}
L. Qu, Y. Zhang, S. Wang, P.-T. Yap, and D. Shen,
``Synthesized 7T MRI from 3T MRI via deep learning in spatial and wavelet domains,''
\textit{Med. Image Anal.}, vol. 62, p. 101663, 2020.

\bibitem{perceptual}
J. Johnson, A. Alahi, and L. Fei-Fei,
``Perceptual losses for real-time style transfer and super-resolution,''
in \textit{Proc. Eur. Conf. Comput. Vis. (ECCV)}, 2016, pp. 694--711.

\bibitem{11523523}
X. Chen, A. Li, Y. Dong, Y. Mao, and Z. Liang,
``Multivariate feature-guided infrared image synthesis with high-frequency fidelity,''
\textit{IEEE Trans. Consum. Electron.}, early access, 2026.

\end{thebibliography}
\end{document}